\documentclass[10pt,twocolumn,letterpaper,amsart]{article}

\usepackage{iccv}
\usepackage{times}
\usepackage{epsfig}
\usepackage{graphicx}
\usepackage{amsmath}
\usepackage{amssymb}
\usepackage{array}
\usepackage{multirow}
\usepackage[utf8]{inputenc}

\usepackage[breaklinks=true,bookmarks=false]{hyperref}

\iccvfinalcopy % *** Uncomment this line for the final submission

\ificcvfinal\fi

\begin{document}

\newcolumntype{L}[1]{>{\raggedright\arraybackslash}p{#1}}

%%%%%%%%% TITLE
\title{Improving Fairness in Large-Scale Object Recognition by CrowdSourced Demographic Information}

\author{Zu Kim \quad Andr\'{e} Araujo \quad Bingyi Cao \quad Cam Askew \\ Jack Sim~\thanks{Currently with Waymo Llc.} \quad Mike Green \quad N'Mah Fodiatu Yilla \quad Tobias Weyand\\
Google Research\\
{\tt\small \{zkim,andrearaujo,bingyi,askewc,jacksim,greenmike,nyilla,weyand\}@google.com}
}

\maketitle
\ificcvfinal\thispagestyle{empty}\fi

\newcommand{\andre} [1]{{\color{magenta}#1}}
\newcommand{\zkim} [1]{{\color{cyan}#1}}

%%%%%%%%% ABSTRACT
\begin{abstract}
There has been increasing awareness of ethical issues in machine learning, and fairness has become an important research topic. Most fairness efforts in computer vision have been focused on human sensing applications and preventing discrimination by people's physical attributes such as race, skin color or age by increasing visual representation for particular demographic groups. We argue that ML fairness efforts should extend to object recognition as well. Buildings, artwork, food and clothing are examples of the objects that define human culture. Representing these objects fairly in machine learning datasets will lead to models that are less biased towards a particular culture and more inclusive of different traditions and values. There exist many research datasets for object recognition, but they have not carefully considered which classes should be included, or how much training data should be collected per class. To address this, we propose a simple and general approach, based on crowdsourcing the demographic composition of the contributors: we define fair relevance scores, estimate them, and assign them to each class. We showcase its application to the landmark recognition domain, presenting a detailed analysis and the final fairer landmark rankings. We present analysis which leads to a much fairer coverage of the world compared to existing datasets. The evaluation dataset was used for the 2021 Google Landmark Challenges, which was the first of a kind with an emphasis on fairness in generic object recognition.
\end{abstract}

\maketitle

%%%%%%%%% BODY TEXT
\section{Introduction}

There has been a significant improvement in machine learning in the last decade, and a good part of it is due to the availability of data at scale for training. A majority of these datasets are collected from the internet. One challenge of using these datasets is fairness. These datasets are created by the users around the world who do not necessarily represent the world’s population properly due to a bias toward a certain group of users with ready access to the internet and smartphones \cite{Jo_2020}. As a result, these biased datasets are used to create recognition systems that are also biased~\cite{mehrabi2019survey}. In this paper, we propose a general approach to reduce bias in such datasets.

There has been increasing awareness on ethical issues in machine learning, and fairness has become an important research topic. It is a new research area, and, so far, most of the efforts have been focused around human sensing (e.g., face recognition) on how different underrepresented groups are fairly represented in the dataset and the training process \cite{pmlr-v80-celis18a,xu2018fairgan,merler2019diversity,Quadrianto_2019_CVPR,yang2020imagenetfairness}. However, there are many more recognition problems that require fairness, and the importance of the dataset’s fairness quality has been undermined \cite{denton2020bringing,sambasivan2021sigchi}. Most relevant efforts here are limited to bias analysis \cite{mehrabi2019survey,weyand2020gldv2} and providing general guidelines on dataset creation \cite{hutchinson2021accountability}. In this paper, we address a fairness issue in general recognition problems that deals with many classes (or a {\em large-scale recognition problem}), where the main fairness question is {\em not} how underrepresented groups themselves are fairly represented in the dataset but how different underrepresented groups’ {\em contributions} are fairly represented in the dataset.

\begin{figure}[t]
\centering
\includegraphics[height=2.2cm]{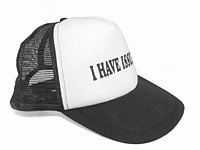}
\includegraphics[height=2.2cm]{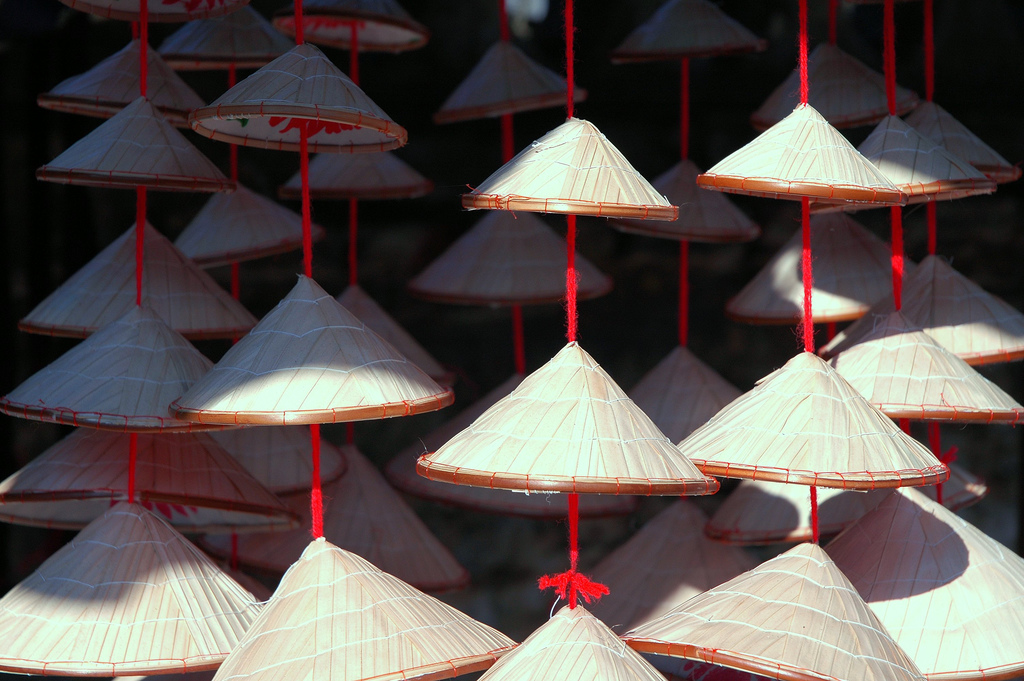}
\includegraphics[height=2.2cm]{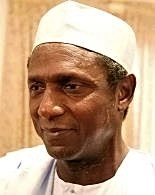}
\caption{Should we include baseball caps, Asian conical hats, or kufis to the recognition classes? There haven't been many (if any) efforts to find a systematic approach to such important decision making.
Photo attributions, left to right:
\href{https://commons.wikimedia.org/wiki/File:Truckerhat.jpg}{1} by Iainf, \href{https://creativecommons.org/licenses/by/2.5/deed.en}{CC-BY},
\href{https://commons.wikimedia.org/wiki/File:N\%C3\%B3n_l\%C3\%A1_\%C4\%91\%E1\%BB\%93_ch\%C6\%A1i.jpg}{2} by terence, \href{https://creativecommons.org/licenses/by/2.0/deed.en}{CC-BY},
\href{https://commons.wikimedia.org/wiki/File:Umaru_Yar\%27Adua_VOA.jpg}{3} by S. Simpson, VOA Photo, Public Domain
}
\label{fig:hats}
\end{figure}

A typical approach in large-scale recognition is first to determine the classes to recognize, collect data for each class, then create a test or training dataset either by equally sampling the data for each class or simply by randomly dividing them.
One hidden challenge here is to determine the classes to recognize. In many cases, this is not even considered an issue (thus, it’s a hidden challenge), and the researchers and developers often use top-N classes based on the dataset availability. However, bias can easily be introduced in that decision. How would you determine whether to add baseball caps or Asian conical hats or kufis to your object detector? In many cases, there is no good justification of choosing one or the other (Figure~\ref{fig:hats}).

\begin{figure}
\centering
\includegraphics[width=8cm]{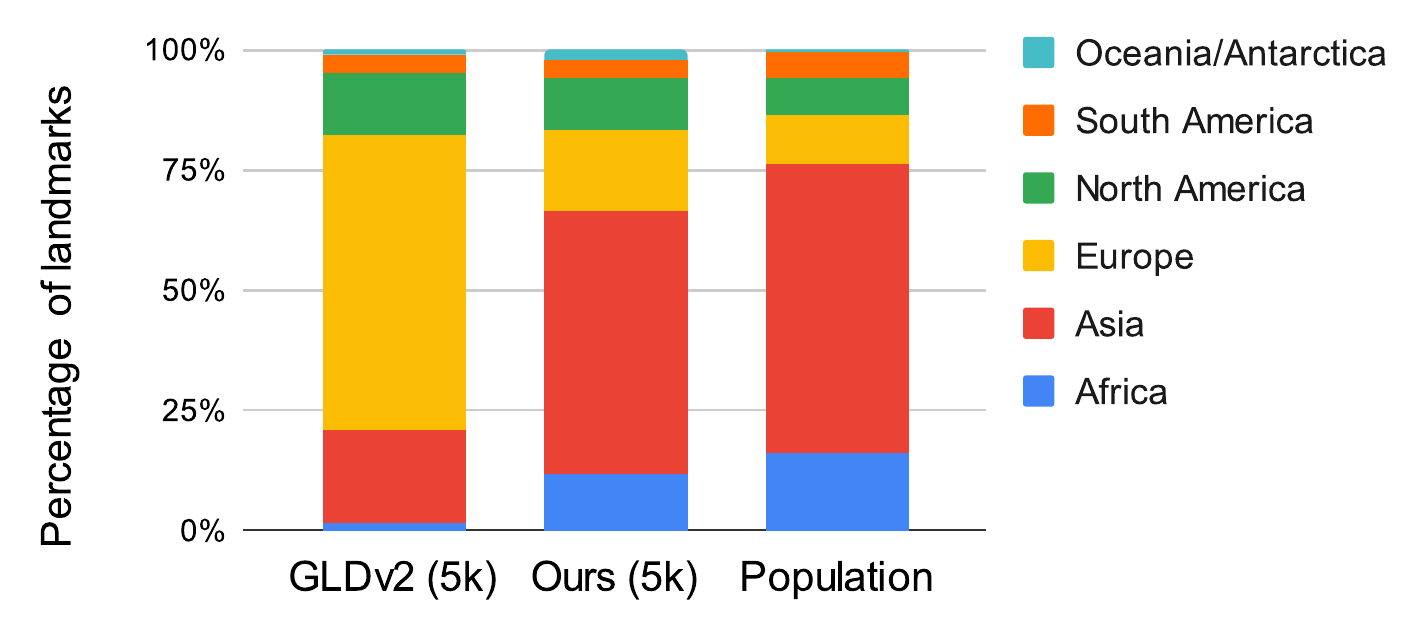}
\caption{Proportion of top-5000 landmarks from each continent available in different datasets (left, center), compared to the world population distribution (right). The Google Landmarks Dataset v2~\cite{weyand2020gldv2} (GLDv2, left) contains significant biases and does not capture the world population fairly.
Our relevance scores (center), constructed using a proposed stratified approach, reflects much better the world population.
The top-5000 landmarks were chosen by the number of images in the dataset (GLDv2) or the relevance scores (ours).}
% \caption{Continent breakdown of the top-5000 landmarks (GLDv2~\cite{weyand2020gldv2} vs ours).}
\label{fig:continent-breakdown}
\end{figure}

Once the classes are determined, a dataset is created by collecting data (e.g., images) from the internet or some other crowdsourcing means. Typically, the number of collected data per class is unbalanced. A commonly-used design decision is to artificially select a similar number of images per class, irrespective of their original distributions (e.g., Caltech-101~\cite{feifei2004caltech101} and ILSVRC~\cite{russakovsky2015imagenet}). In these cases, often the number of test images is held constant for all classes, while the number of training images does not vary significantly. For a long-tailed recognition problem, a typical strategy is to collect all the available data (e.g., iNaturalist~\cite{horn2018inat} and Google Landmarks Dataset v2~\cite{weyand2020gldv2}) and to use the distribution of these available data (or uniform distribution in \cite{iscen2021}) for training and testing. There is no good justification for either of them. At a glance, the former seems to be fairer, but that’s not necessarily the case, especially for large-scale (or long-tail) recognition problems. On the practical side, there is not much data available for the classes that belong to the long tail, but, more importantly, {\em not all classes are equally important (relevant)}. For example, if there is a landmark recognition model, most people would expect it to recognize (and well-recognize) the Eiffel Tower but only a smaller number of users would care about lesser-known neighborhood landmarks. Some classes are essential and some others are nice-to-have.

In many of the large scale recognition problems, the relevance of a class (to the recognition objective) is implicitly determined and used for either class selection or training (or data collection), and the resulting dataset would contain bias \cite{mehrabi2019survey, weyand2020gldv2}. Our focus in this paper is on the process of providing a fair definition(s) of relevance, and dealing with biases in estimating the relevance in crowdsourced data.

To our knowledge, this work is the first one to suggest a general approach to improve fairness in datasets for general (image) recognition. Furthermore, this is the first attempt to explicitly assign and estimate relevance to each of the classes we are trying to recognize. We present our approach around this and its application to the landmark recognition domain to prove its necessity and effectiveness. We provide in-depth analysis on how crowdsourced data can be biased and how different demographic groups can have different relevance to each of the classes. Finally, we present a method to construct a less-biased landmark dataset for evaluation with a baseline experimental result.

%------------------------------------------------------------------------
\section{Approach}

Our approach starts with defining the {\em relevance} of each class to the recognition objective. The recognition objective should be determined based on the project usage. For example, a landmark recognition system that would be used for a navigation system would have different objectives from the one that is used for tourism applications.

In most cases, recognition systems are for human users. Therefore, to build a useful and fair object detector, one needs to define {\em relevance} in a way to fairly represent all populations. In fact, relevance is a very personal term, and everyone has their own relevance criteria. Therefore, one way to define the relevance of a class is to use an aggregation: fair collective opinions. In addition, there can be more than one objective and definitions of relevance. For further discussion, see our definitions of landmarkiness in the following section.

\begin{figure}
\centering
\includegraphics[width=6cm]{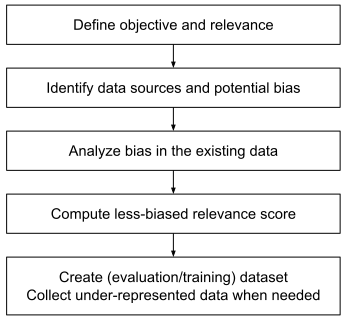}
\caption{Proposed approach to create fairer dataset for large scale object recognition}
\label{fig:approach}
\end{figure}

Figure~\ref{fig:approach} illustrates our basic approach to creating a fairer dataset for large-scale object recognition. Once the objective and the relevance is defined, we need to identify the data sources and potential sources of biases for each data source. We focus on crowdsourced data such as Wikipedia and commercial services such as photo collections or maps that allow user contributed content. In crowdsourced data, the relevance is strongly tied to the amount of information that the users contribute. In fact, this is an implicit assumption of most of the existing data collection practices as discussed in the introduction. While we agree that the amount of user contribution is a strong proxy for relevance, we need to reduce potential biases that come with it. We present an approach to reduce such biases.

Our approach is to use demographic information of the contributors, such as the age, gender, and the country of origin, which is available in many of the crowdsourcing systems as basic login information. By using this available information, we can analyze the bias in the raw data, and using this analysis result, we can come up with a {\em fairer relevance score} for each of the classes. We use stratification to compute the relevance scores. See the next subsection for the detailed method.

Then we can use these relevance scores to re-balance the evaluation or training dataset. One may use the probability sampling to re-create a fairer subset of the full dataset, or use the full dataset with involving the relevance scores in the training process (e.g., as a part of the loss function) because more data is better in general in machine learning systems. It is not in the scope of this paper to suggest a good training approach. One potential challenge is an absolute lack of under-represented data, and we may need to find a way to collect more data from underrepresented demographic groups such as by providing incentives.

Finally, there are several important considerations of applying the proposed approach to a real problem: 
\begin{itemize}
\item The available demographic information is not likely to be sufficient to address all the biases. After all, crowdsourced data as a whole only represents the population that can access the internet. Variables such as country of origin or income level (if available) can relieve this bias but we have to acknowledge the fundamental limitation of the approach. We need to carefully review the limitations and transparently inform the users.
\item The demographic information is highly private information, and we need to handle it with extra care. For example, if there is a class where its user content is uploaded by a single user, the analysis may reveal the user’s demographic information. Therefore, the data visibility and the aggregation level will have to be carefully reviewed. Adding differentially private noise \cite{dwork_2008_differential_privacy} to the result can also be an option to consider.
\item 1:1 representation of the population can potentially suppress the minority voice. We need a measure to make sure that an underrepresented demographic group is not penalized by the popular vote. We add this consideration to our relevance score formulation in the following subsection, so that the data curators can choose the degree of underrepresented groups' representation.
\end{itemize}

\subsection{Relevance score generation based on contributor’s demographics}
\label{sec:formulation}

In statistics, stratified sampling~\cite{Neyman_1992_Stratified_Sampling} has been used to make survey results better represent the entire population. A similar approach can be used to compute relevance scores of crowdsourced data. The degree of fair relevance can be proxied by the expected amount of crowdsourced data (e.g., the number of photos uploaded) if everyone in the world {\em would have} fairly contributed to the data. We use the following probability equation to estimate the (ideal) distribution of the classes in a fairly contributed dataset:
\begin{equation} \label{eq:raw_fomulation}
\begin{aligned}
P(c) &= \sum_h{P(c|h) P(h)} \\
     &= \sum_h{P(c|h_1,h_2,\ldots) P(h_1,h_2,\ldots)},
\end{aligned}
\end{equation}
where $P(c)$ is the probability distribution over the class, $H_1, H_2, \ldots$ represent demographic categories, $h = h_1, h_2, \ldots$ represents a specific demographic group (strata), and $P(h)$ is a shorthand for $P(H=h)$. For example, $h$ represents Canadian females, and $h_1$ is for females ($H_1$ is for gender), and $h_2$ for Canadian residents ($H_2$ is for country of residence). $P(c|h)$ is the distribution of the contribution from people who belong to the demographic group $h$. This distribution can be estimated from the actual contribution statistics of the contributors of the given demographic group:

\[
P(c|h) = \frac{A_{c,h}}{A_{h}},
\]
Where $A_{c,h}$ is the amount of contributions of the demographic group $h$ to the class $c$ and $A_{h}$ is the total amount of contributions of the demographic group $h$ to all the classes.

Naive crowdsourcing with no bias adjustment essentially assumes that the prior probability of a demographic group, $P(h)$, follows the contributors’ demographic distribution. Apparently, using the world’s population statistics is a much better (and fairer) way to estimate $P(h)$. The demographics information can be obtained from various sources that provide demographic statistics including \cite{PopulationPyramid, DataCommons}. For example, a prior for Candian ($h_1$) females ($h_2$),
\[
P(h_1, h_2) = \frac{N_{h_1,h_2}}{N_W},
\]
where $N_{h_1,h_2}$ is the Canadian female population and $N_W$ is the World’s population. Note that we may not have access to the fine demographic category (e.g., the population of women in their 30's living in Montreal), in which case we can approximate it by assuming independence (or conditional independence) among the demographic groups. For example, $P(h_1, h_2) = P(h_1) P(h_2)$.

For some demographic groups (e.g., non-binary gender) there may not be sufficient population statistics to infer $P(h)$. In such a case, one may fall back to use the contributor statistics to estimate the prior. For example:
\[
P(H=h_a) \approx \frac{M_a}{M_W},
\]
where $M_a$ is the number of contributors that belong to the demographic group $h_a$ and $M_W$ is the total number of contributors. In this case, we also need to adjust the prior (estimated from population statistics) of other demographic groups in the same category:
\begin{align*}
P(h_{x, x \neq a}) \approx \frac{N_x}{N_W} (1 - P(h_a)),
\end{align*}
where $N_x$ is the population of the demographic group $h_x$.
We can also handle mixtures of categories (e.g., non-binary gender living in Canada) by assuming independence or conditional independence. Note that because we are using the contributor distribution to guess the prior, this unavailable demographic group may be underrepresented. To mitigate the issue one may utilize the diversity boost term we introduce later in this section to improve their representation.
Another note is that we need to distinguish this demographic group (with unavailable population) from an unspecified group. For example, the contributors who chose to specify their gender as `other' are different from the contributors who did not specify their gender due to various reasons. Depending on the use case, one may choose to apply the above equation to this unspecified group or ignore their data assuming that their actual distribution follows the other contributors' distribution. For example, depending on the registration process, unspecified gender may or may not mean non-binary gender but a contributor of unspecified age is likely to follow the general contributor statistics, and we can ignore them (otherwise, this data will undermine underrepresented groups).

\begin{figure}[t]
\centering
\includegraphics[width=6cm]{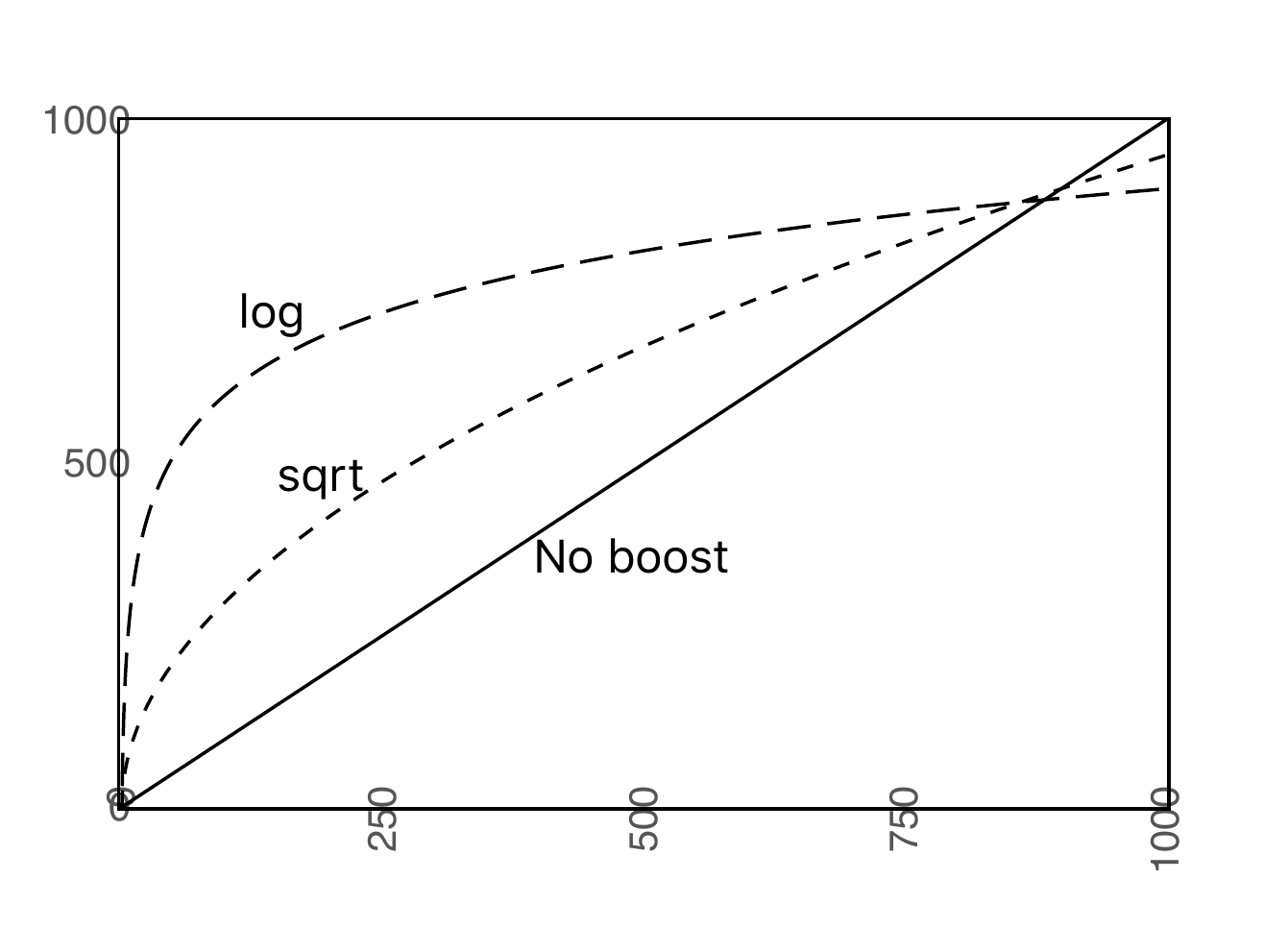}
\caption{Example {\em diversity boost functions} to increase the representation of a minority group.}
\label{fig:boost_functions}
\end{figure}

As mentioned in the previous section, we would like to introduce a term to increase the representation of a minority demographic group. In this case, what we are estimating is rather an expected utility \cite{schoemaker1980expectedutility}. We can add a diversity boost term to the equation~\ref{eq:raw_fomulation}:
\[
U(c) = \sum_h{\alpha_h P(c|h) P(h)},
\]
where $U$ is the expected utility of the class and $\alpha_h$ is the {\em diversity boost} term. The diversity boost term can be any function of the demographic group $h$. For example, if it is an inverse of $P(h)$ all the demographic groups would be treated equally regardless of its population. For more intuitive formulation, we introduce a growth function with respect to the population ({\em diversity boost function}):
\[
\alpha_h = \frac{f(N_h)}{N_h},
\]
where $f(x)$ is the {\em diversity boost function}. Note that $P(h)$ is proportional to $N_h$. Therefore, if we use $f(x) = sqrt(x)$ or $log(x+1)$, for example, $\alpha_h P(h)$ will grow slower than the actual population (Figure~\ref{fig:boost_functions}), and lower population groups would be better represented. It is up to the data curator to choose a proper diversity boost function based on the use case.

The data curator also needs to carefully choose the level of the demographic hierarchy (e.g., continent to country or coarse age groups to fine age groups) where the diversity boost function would be applied to. Applying a diversity boost function (unless it’s linear) to different levels will give different results, i.e., $f(x + y) \neq f(x) + f(y)$.
Another note is that one may choose to apply the diversity boost term only to certain demographic groups (e.g., to the country but not to the age). For example,
\[
\alpha_{h_1, h_2} = \frac{f(N_{h_1})}{N_{h_1}},
\]
where $h_1$ represents the country of living and $h_2$ is the age group.
It is also possible to apply them independently
\[
\alpha_{h_1, h_2} = \frac{f_1(N_{h_1})}{N_{h_1}} \frac{f_2(N_{h_2})}{N_{h_2}},
\]
if $P(h_1, h_2)$ was estimated based on the independence assumption.

%------------------------------------------------------------------------
\section{Fairness in landmark recognition}

Landmark recognition is an active research area \cite{zheng2009tour, avrithis2010retrieving, Weyand2015, noh2017large}. However, fairness in landmark recognition has rarely been touched. The most relevant work is \cite{weyand2020gldv2}, where Weyand et al. present the geographical distribution of their dataset (Figures 3 and 4) showing that the landmarks in some parts of the world, for example, Africa, India, and China, are significantly under-represented compared with their population. It illustrates how much bias can be introduced by plain collection of crowdsourced data. However, we do not think that the relevance of the landmark simply depends on where they are (or what the population is near the landmark), and we cannot achieve fairness just by balancing out the number of landmarks per country or continent. In this section, we present how we apply the proposed approach to landmark recognition to create a less-biased landmark recognition dataset.

%------------------------------------------------------------------------
\subsection{Fair definitions of landmarkiness}

Following our suggested approach, we first define the fair landmark recognition objectives and relevance (`landmarkiness') of each landmark. One of our findings is that there are multiple usages of a landmark recognition system with varying objectives. Among them, we are interested in the definitions stated in Table~\ref{table:fair-landmarkiness}. Note that these definitions are highly personal, and we are interested in fairer aggregations of these personal relevance criteria.

\begin{table*}
 \begin{center}
 \begin{tabular}{|L{4cm}|L{10cm}|}
 \hline
 Recognition objective & Fair definition of relevance \\
 \hline
 Personal Importance / Utility &
 “Which places are personally important (or useful)? (e.g., grocery market, school, local social  security office)” \\
 \hline
 Tourism (Personal) &
 “If everyone in the world can go anywhere in the world, what is the place they would like to go/remember?” \\
 \hline
 Tourism (Cultural and Historical Importance) &
 “Which place is culturally and historically important to each group of people (e.g., local/ethnic community or country)?” \\
 \hline
 \end{tabular}
 \end{center}
 \caption{Fair definitions of landmarkiness we are interested in. Our goal is to get fairer aggregations of these personal relevance.}
 \label{table:fair-landmarkiness}
\end{table*}

%------------------------------------------------------------------------
\subsection{Data sources, bias, and limitations}

There are multiple sources to estimate the relevance other than crowdsourcing. For example, one could infer tourism scores from tour guide websites (e.g., published by local governments) and cultural and historical relevance from organizations such as Unesco. However, collecting such data is difficult to automate and they too contain various sources of biases such as tech ability or funding of a local government.

We propose to infer a personal tourism score using online map (Google Maps) image upload statistics. The online map image uploads reflect both personal and touristic relevance. We expect the plain statistics to be biased (in a similar way to that of \cite{weyand2020gldv2}) because the uploaders are not likely to represent the general population. For example, there is a large variation in internet accessibility and proficiency by place of living (countries, urban or rural), education level, gender, income level, travel ability, and age \cite{Jo_2020}. We cannot capture all these aspects for data stratification, but we can reduce bias in areas where the demographic information is available: gender, age, and country of residence. Strict privacy guidelines were applied so that the individual demographic information is not accessed or inferred during or after the data aggregation.

We first present bias analysis around these three demographic categories, and use that information to create a less-biased landmark recognition dataset.

%------------------------------------------------------------------------
\subsection{Bias analysis of online map image uploads}

In this section, we present our observations on the plain (un-stratified) upload statistics. Our proxy of relevance is the number of images uploaded. Since there can be a wide difference between the number of photo uploads per person (one to say, 1000 or more), we used a log-scale with thresholding in this analysis so that a contributor who uploads a single photo does not get penalized.

We first analyzed the demographic distribution of the image uploaders per continent. One of our notable findings was that African uploaders are highly under-represented. There were many times more European uploaders than African uploaders while the population of Africa is almost twice that of Europe. Female uploaders were under-represented in all continents, and the gap was larger in Asia and Africa. We were able to observe significant under-representation of the older generation (60 or higher) in all continents. The age distribution varied greatly by countries. In some countries, a majority of the contributions was by a younger population (say, under 35), while in some other countries mid-age groups (e.g., 35-60) contributed more.

Table~\ref{table:top10-continent} shows top 10 (based on aggregated log-scale contributions) landmarks by Africans {\em vs.} Europeans. Note that this breakdown is by the uploaders, not the location of the landmark. For example, the Eiffel Tower is the 3rd most uploaded landmark by the African uploaders. We see from the table that the overall ranking over-represents Europeans but under-represents Africans. The difference is not only observed among the continents, but also within a continent. For example, the top 20 African landmarks only consisted of South African, Egyptian (note that both South Africans and Egyptians are already under-represented), European and Asian landmarks, and the top landmark by Nigerians (highest population in Africa) only appears at top 26. One thing to note is that for a majority of the population it is difficult to travel abroad and take photos, and it is difficult to meet a part of our definition “If everyone in the world can go anywhere in the world”. On the other hand, this result respects more of the local landmarks by local people who know the landmarks better than visitors.

\begin{table*}
\begin{center}
\begin{tabular}{|c|L{4cm}|L{4cm}|L{4cm}|}
\hline
Rank & Overall (No stratification) & Africans & Europeans \\
\hline
\hline
1 & Eiffel Tower$^\ast$ (France) &
Table Mountain National Park (S. Africa) &
Eiffel Tower (France) \\
\hline
2 & Colosseum (Italy) &
V\&A Waterfront (S. Africa) &
Colosseum (Italy) \\
\hline
3 & Taj Mahal (India)  &
Eiffel Tower (France) &
St. Peter's Basilica (Vatican City) \\
\hline
4 & Louvre Museum (France) &
Giza Necropolis (Egypt) &
Louvre Museum (France) \\
\hline
5 & St. Peter's Basilica (Vatican City) &
Masjid al-Haram (Saudi Arabia) &
Charles Bridge (Czech Rep.) \\
\hline
6 & Central Park (US) &
uShaka Marine World (S. Africa) &
Red Square (Russia) \\
\hline
7 & Walt Disney World\textsuperscript{\textregistered} Resort (US) &
Al Masjid an Nabawi (Saudi Arabia) &
Trevi Fountain (Italy) \\
\hline
8 & Tower Bridge (UK) &
Salah Al-Din Al-Ayoubi Castle (Egypt) &
Tower Bridge (UK) \\
\hline
9 & Charles Bridge (Czech Rep) &
Citadel of Qaitbay (Egypt) &
Roman Forum (Italy) \\
\hline
10 & Times Square (US) &
Citystars Heliopolis (Egypt) &
Disneyland Paris (France) \\
\hline
\end{tabular}
\end{center}
\caption{Top-10 landmarks by Africans and Europeans. $^\ast$: Also includes the photos uploaded to Champ De Mars.}
\label{table:top10-continent}
\end{table*}

We also analyzed landmark rankings by gender and age. Note that there are high variations of gender and age gaps by continents and countries. For example, the top landmark ranking by the male uploaders contains more Asian landmarks not because the male uploaders prefer Asian landmarks but because the gender gap is larger in Asia than other regions. Therefore, we conditioned the result on the country or continent of living. The differences by gender and age were not as big as those by countries and continents but we were able to observe some trends. For example, we analyzed the landmark rankings by North American demographic groups, and observed that female uploaders tend to upload more images of children’s theme parks while older populations prefer natural parks.

To summarize, we observe that the crowdsourcing contributors are highly unbalanced by gender, age, and place of living, and each of the demographic groups have largely different relevance criteria and interests.

%------------------------------------------------------------------------

\subsection{Less-biased relevance scores}

We followed the formulation in Section~\ref{sec:formulation} to estimate a less-biased relevance score for each landmark. We used the demographic statistics from \cite{PopulationPyramid} which provided the populations by gender and age for most countries. For the countries where such fine-grained data is not available, we combined the country's whole population obtained from \cite{DataCommons} and Wikipedia with the World's gender and age distributions assuming independence between them. 
Table~\ref{table:top10-stratified} shows the top-10 landmarks as a result of the stratification. We can observe the increased diversity compared with the raw result in Table~\ref{table:top10-continent}.

\begin{table*}
\begin{center}
\begin{tabular}{|c|c|c|}
\hline
Rank & Top-10 landmarks (No stratification) & Top-10 landmarks (Stratified) \\
\hline
\hline
1 & Eiffel Tower (France) & Eiffel Tower (France) \\
\hline
2 & Colosseum (Italy) & Taj Mahal (India) \\
\hline
3 & Taj Mahal (India) & Louvre Museum (France) \\
\hline
4 & Louvre Museum (France) & Masjid al-Haram (Saudi Arabia) \\
\hline
5 & St. Peter's Basilica (Vatican City) & Al Masjid an Nabawi (Saudi Arabia) \\
\hline
6 & Central Park (US) & Colosseum (Italy) \\
\hline
7 & Walt Disney World\textsuperscript{\textregistered} Resort (US) & The Dubai Mall (Arab Emirates) \\
\hline
8 & Tower Bridge (UK) & National Mall (US) \\
\hline
9 & Charles Bridge (Czech Rep) & Giza Necropolis (Egypt) \\
\hline
10 & Times Square (US) & Trocadéro Gardens (France) \\
\hline
\end{tabular}
\end{center}
\caption{Top-10 landmarks before and after stratification.}
\label{table:top10-stratified}
\end{table*}

\begin{figure}
\centering
\includegraphics[width=8cm]{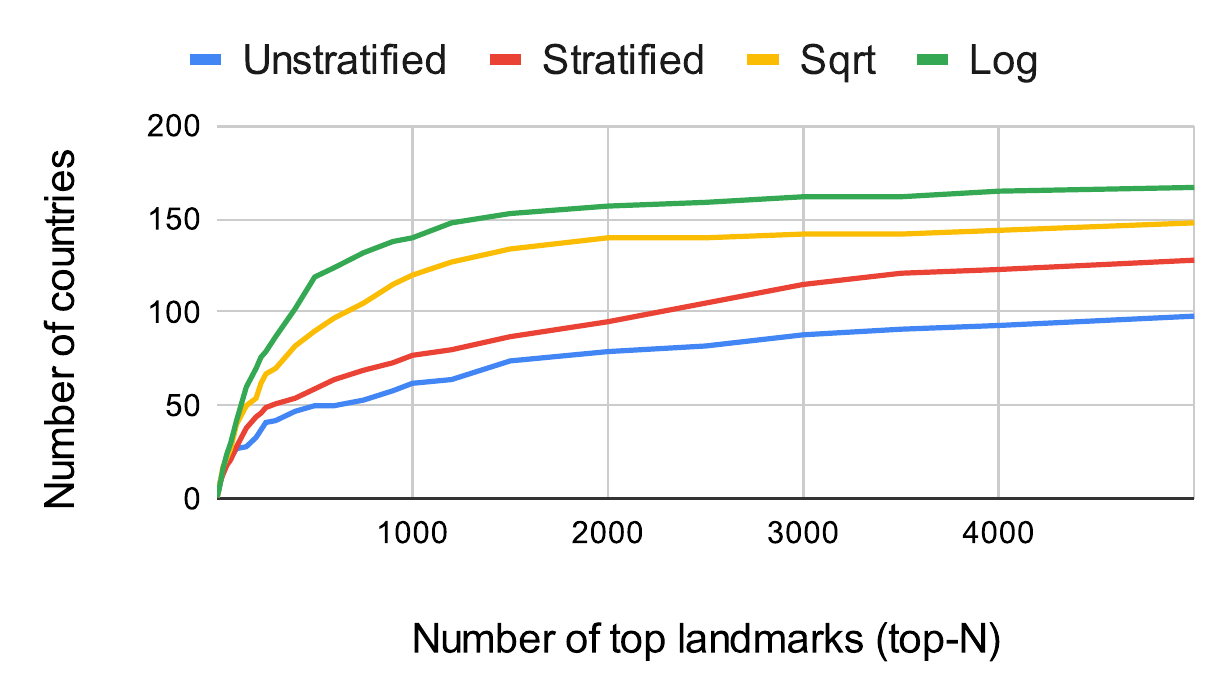}
\caption{Number of countries that host the top-N landmarks. We see that the stratification (red) increases diverse representation, and the diversity boost functions further enhance the diversity.}
\label{fig:country-breakdown}
\end{figure}

The effect of the stratification and diversity boosting is shown in Figure~\ref{fig:country-breakdown}. We applied $sqrt$ and $log$ diversity boost functions on the country of living, and computed the ranking. We can clearly see that the stratification helps increase diverse representation, and the diversity boost functions further enhance the diversity.

We further compare our result with the Google Landmarks Dataset v2~\cite{weyand2020gldv2} (GLDv2). The continent breakdown of the two datasets is shown in Figure~\ref{fig:continent-breakdown}, where we can also observe the enhanced diversity.

One important finding is that only 65\% of the new top-10000 landmarks (computed with pure stratification) overlap with the top-10000 un-stratified landmarks. In other words, if one uses plain upload statistics to build a 10000-landmark classifier, 35\% of the under-represented landmarks would be omitted from it.

\subsection{Fairer evaluation dataset}

We used the relevance scores to create fairer evaluation datasets and performed a baseline experiment.

We applied random probabilistic sampling based on the relevance scores for determining the number of samples per landmark. An example application is shown in Table~\ref{table:sampling_example}. By applying the probabilistic sampling, long tail examples would not be omitted from the evaluation dataset.
The images were downloaded from Wikimedia.

The details of the evaluation data construction is not a main topic of this paper, and we present that in \cite{kim2021gld}. We also present the details of the datasets (e.g., where to download and the limitations) in that technical report. The new evaluation datasets change the baseline performance as expected, and the new baseline is shown in Table~\ref{table:evaluation}.

\begin{table*}
    \centering
    \begin{tabular}{c|c|c}
        \hline
        Landmark name & Relevance score & \# of images would be sampled \\
        \hline
        Giza Necropolis (Egypt) & $\sim 0.000827$ & 28 \\
        Srikalahasti Temple (India) & $\sim 0.000070$ & 0 \\
        Acuario de Veracruz (México) & $\sim 0.000056$ & 3 \\
        \hline
    \end{tabular}
    \caption{Example relevance scores and numbers of the images that would be added to the evaluation dataset by the probabilistic sampling. By using probabilistic sampling, we can preserve the long tail examples.}
    \label{table:sampling_example}
\end{table*}

\begin{table*}
\begin{center}
\begin{tabular}{c|c|c}
\hline
Challenge & Old baseline & New baseline \\
\hline
Retrieval challenge (mAP@100) & 0.2258 & 0.2153 \\
Recognition challenge ($\mu$AP@100) & 0.2293 & 0.2376 \\
\hline
\end{tabular}
\end{center}
\caption{Baseline results changes with the new dataset. DELG global embedding~\cite{cao2020delg} was used for the evaluation.}
\label{table:evaluation}
\end{table*}

Finally, we present an experimental result showing that our evaluation dataset effectively discriminates carefully designed classifiers with fairness in mind from conventional classifiers.
We compared the results with the classifiers built with and without using the relevance scores. The first (more biased) classifier uses a $\sim 84$k randomly (uniform distribution) chosen subset from GLD’s {\em training dataset}, and the second (less-biased) classifier uses another random $\sim 84$k subset created using the distribution of relevance scores. We made sure that all the landmarks in the evaluation datasets were present in both of the $\sim 84$k subsets.

\begin{figure*}
\centering
\includegraphics[width=10cm]{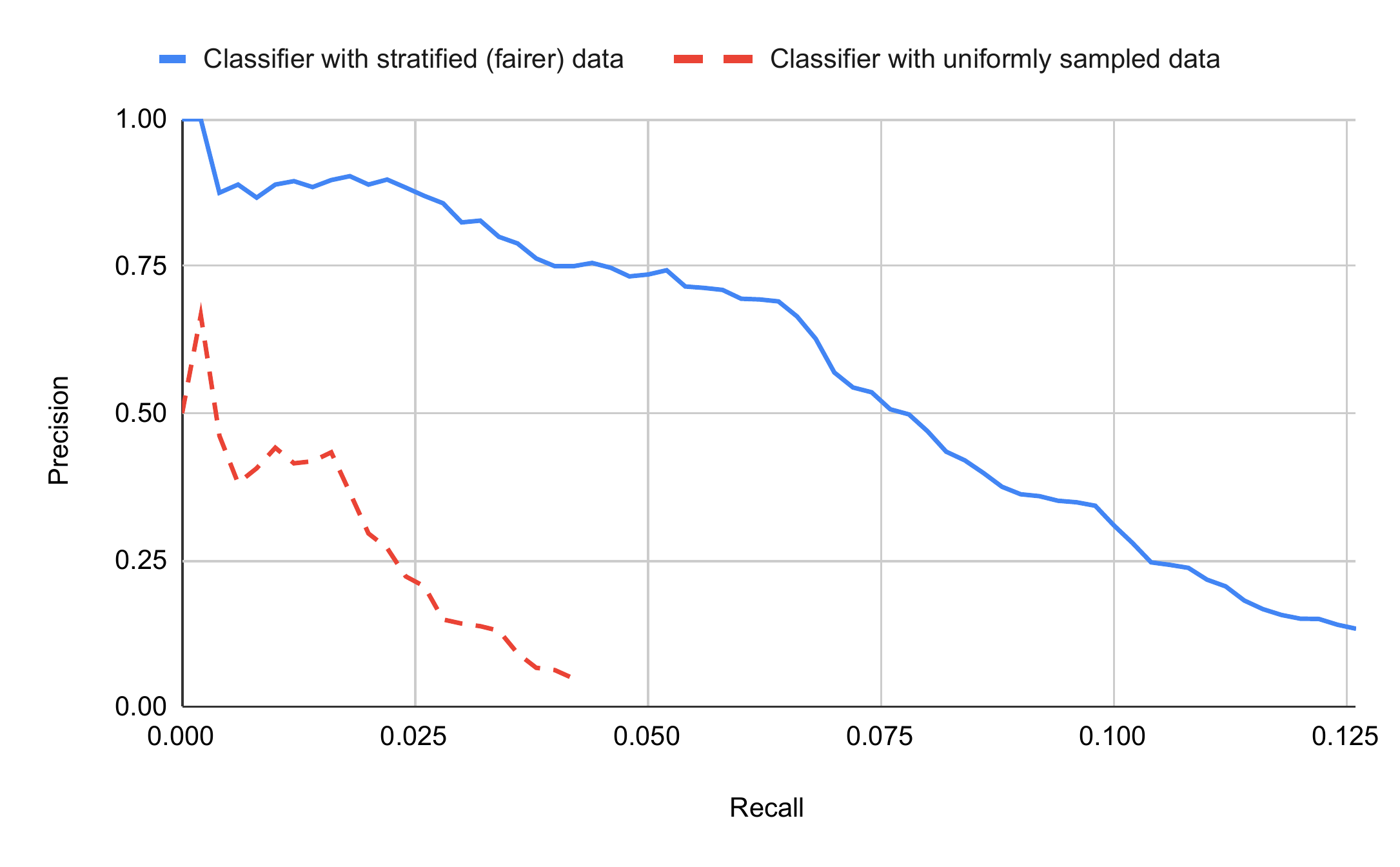}
\caption{Baseline result (top-1 precision vs recall) on the uniformly sampled training data vs stratified (fairer) training data.}
\label{fig:baseline-result}
\end{figure*}

The classifier we use is based on $k$-nearest neighbors retrieval. We used a DELG embedding extractor \cite{cao2020delg} trained with GLDv2 train-clean dataset \cite{yokoo2020twostage} with the training datasets (the 84k subsets) as indexes. The result (top-1 precision vs recall) is presented in Figure~\ref{fig:baseline-result}. We can see that the new evaluation dataset can clearly discriminate between the classifiers trained with and without fairness considerations.

The new evaluation dataset was used in the 2021 Google Landmark Recognition~\cite{glrecognition2021} and Retrieval~\cite{glretrieval2021} competitions with over 300 participating teams. In this challenge, we have not provided the relevance scores to the participants. Yet, one of the winning teams~\cite{yuqi2021gld2nd} explicitly formulated fairness by their own (continent-aware sampling strategy and landmark-country aware reranking), and proved that it improved their performance.

%------------------------------------------------------------------------
\section{Conclusions and future work}

We introduced a new paradigm for large-scale object recognition, where we define and estimate the {\em relevance} of each class in the training data.
Implicit assumptions (or non-assumptions) on the relevance used in the conventional machine learning practices introduce biases to the datasets and the resulting models.
We presented an approach to explicitly define the relevance and estimate them in a fairer manner by using the demographic information of the crowdsourcing contributors. 
We applied the approach to landmark recognition, and demonstrated its effectiveness through analysis and experiments.
We are planning to apply the suggested approach to another domain, such as in creating a generic object recognition dataset.

{\small
\bibliographystyle{ieee_fullname}
\bibliography{egbib}
}

\end{document}